\documentclass[letterpaper, 10 pt, conference]{ieeeconf}
\IEEEoverridecommandlockouts 

\usepackage{amsmath}
\usepackage{graphicx}
\usepackage{amsfonts}
\usepackage{subcaption}
\usepackage{cite}
\usepackage{verbatim}
\usepackage{float}
\usepackage{array}
\usepackage{authblk}
\usepackage{amssymb}
\usepackage{multirow}
\usepackage[letterpaper, left=48pt, right=48pt, bottom=43pt, top=57pt]{geometry}
\usepackage{hyperref}
\usepackage{svg}
{}%
\newcolumntype{P}[1]{>{\centering\arraybackslash}p{#1}}
\usepackage{algpseudocode}
\usepackage{algorithm}
\usepackage{nicefrac}

\title{A Learned Stereo Depth System for Robotic Manipulation in Homes}
\author{Krishna Shankar, Mark Tjersland, Jeremy Ma\thanks{Corresponding author: jeremy.ma@tri.global}, Kevin Stone, Max Bajracharya \\ Toyota Research Institute, Los Altos, CA \thanks{Author contacts: krishna@polycam.ai, \{first.last\}@tri.global}}

\graphicspath{{figure/}}
\begin{document}
\maketitle
\thispagestyle{empty}
\pagestyle{empty}
\bstctlcite{IEEEexample:BSTcontrol}
\begin{abstract}
  We present a passive stereo depth system that produces dense and accurate point clouds optimized for human environments, including dark, textureless, thin, reflective and specular surfaces and objects, at 2560x2048 resolution, with 384 disparities, in 30 ms.  
  The system consists of an algorithm combining learned stereo matching with engineered filtering, a training and data-mixing methodology, and a sensor hardware design.
Our architecture is 15x faster than approaches that perform similarly on the Middlebury and Flying Things Stereo Benchmarks.
To effectively supervise the training of this model, we combine real data labelled using off-the-shelf depth sensors, as well as a number of different rendered, simulated labeled datasets. 
We demonstrate the efficacy of our system by presenting a large number of qualitative results in the form of depth maps and point-clouds, experiments validating the metric accuracy of our system and comparisons to other sensors on challenging objects and scenes. We also show the competitiveness of our algorithm compared to state-of-the-art learned models using the Middlebury and FlyingThings datasets.
\end{abstract}
\section{Introduction}

High resolution, high frame rate, low power, accurate depth information is critical for performing robotic tasks including for mapping, scene understanding, collision avoidance, and manipulation.  Most existing sensors and algorithms such as LIDAR \cite{velodyne_lidar_2018}, passive hand coded dense stereo matching \cite{schauwecker2015sp1}, \cite{scharstein2002taxonomy}, and structured light sensors\cite{bamji2018impixel} fail on different surface types, such as shiny, specular, dark, textureless, thin, or transparent materials, which are particularly common in human environments.  For a robot to successfully operate in these environments our objectives for depth sensing are:


\begin{itemize}
    \item Being able to see all or most of the robot's manipulation workspace at once with sufficient spatial resolution, without moving the head or adding multiple sensors. This simplifies the overall robot system architecture, and better performance in the case of dynamic scenes.
    \item Producing dense depth maps on low-texture, dark, reflective and transparent objects and surfaces.
    \item Operating at 5-10Hz frame rate.
    \item Maintaining a level of depth accuracy in the workspace that enables accurately forming a $2$ cm voxel map. This is driven by ensuring we can avoid collisions in the workspace robustly and manipulate the smallest object our gripper can grasp.
\end{itemize}

To achieve these objectives we have developed a stereo-depth system optimized for use in homes at manipulating range. Our contributions include:
\begin{itemize}
    \item Developing an efficient learned stereo-matching algorithm that builds on prior work with a number of critical optimizations, including a post-processing step that combines a learned confidence measure along with classical filtering to ensure metrically accurate point clouds. Notably, this algorithm runs much faster than those prior in terms of milliseconds/megapixel, scaling well to high resolutions by leveraging GPU parallelism. 
    \item Using a mix of real and synthetic data, with real data from our sensor head labelled using off-the-shelf sensors, and a blend of synthetic data spanning a range of qualities to produce accurate depth-maps on challenging surfaces and objects.
    \item Designing a sensor-head optimized for the (large) workspace of the robot, and enabling high depth accuracy for collision avoidance and manipulating smaller objects (as small as 2cm). 
\end{itemize}


\begin{figure}[!t]
     \begin{subfigure}[b]{\textwidth}
         \includegraphics[width=0.49\textwidth]{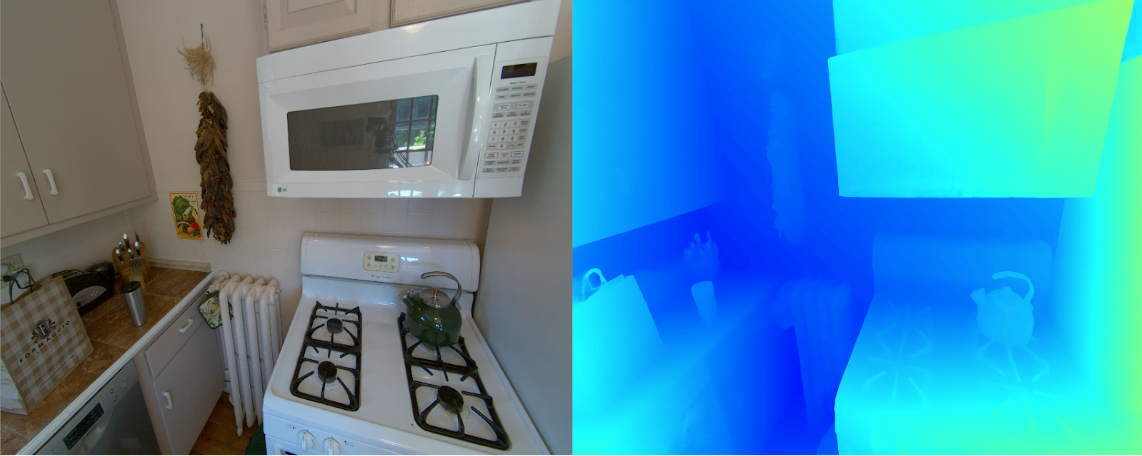}
         \label{fig:y equals x}
     \end{subfigure}
     \begin{subfigure}[b]{\textwidth}
         \includegraphics[width=0.49\textwidth]{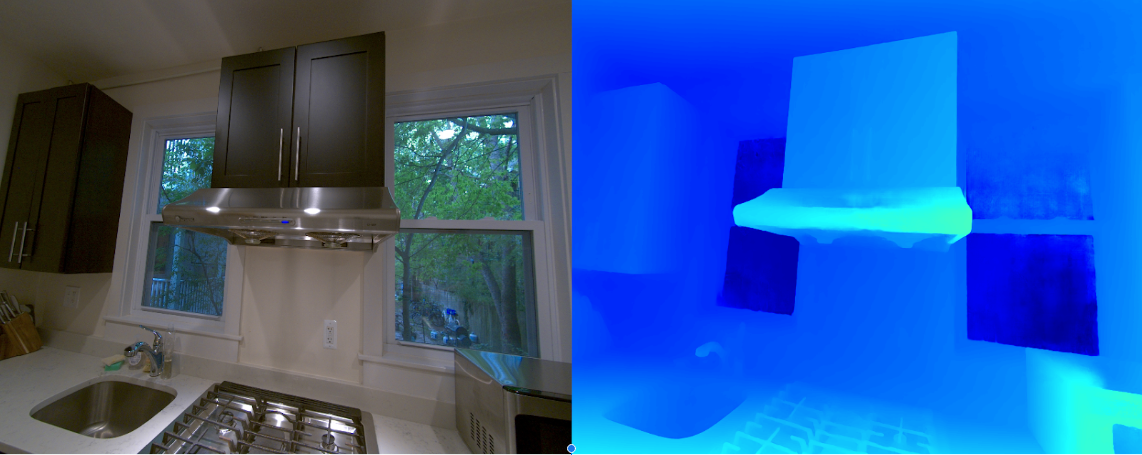}
         \label{fig:three sin x}
     \end{subfigure}
        \caption{Corresponding color and depth images for difficult scenes in real homes.}
        \label{fig:intro_image}
\end{figure}
\begin{figure*}[!htb]
     \centering
     \begin{subfigure}[b]{0.29\textwidth}
         \centering
         \includegraphics[width=\textwidth]{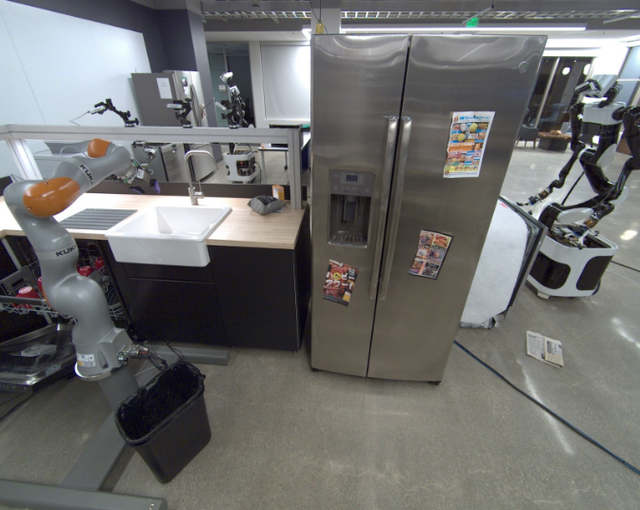}
         \caption{RGB image of brushed metal fridge and floor with reflections.}
         \label{fig:fridge_rgb}
     \end{subfigure}
     \hfill
     \begin{subfigure}[b]{0.29\textwidth}
         \centering
         \includegraphics[width=\textwidth]{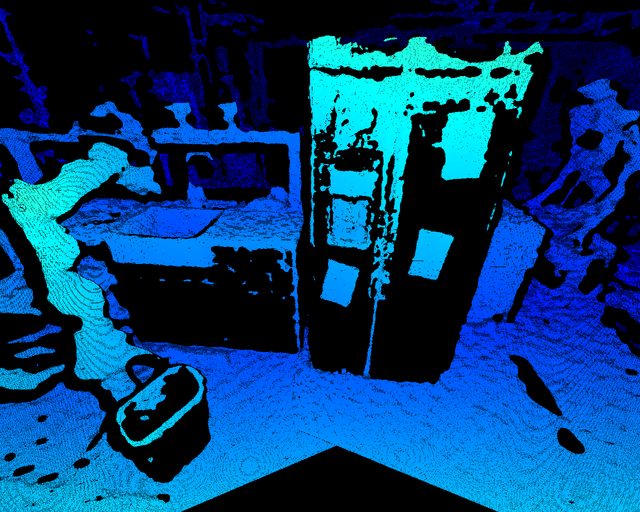}
         \caption{Intel Realsense projected-texture stereo depth coverage.}
         \label{fig:fridge_realsense}
     \end{subfigure}
     \hfill
     \begin{subfigure}[b]{0.29\textwidth}
         \centering
         \includegraphics[width=\textwidth]{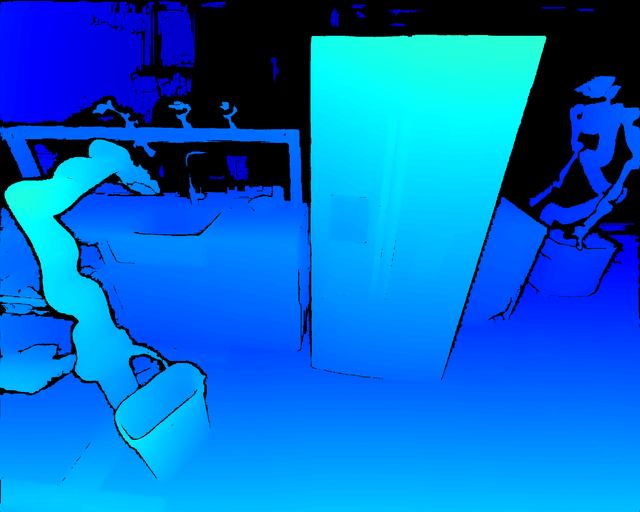}
         \caption{The output depth map of our algorithm.}
         \label{fig:five over x}
     \end{subfigure}
     \begin{subfigure}[b]{0.29\textwidth}
         \centering
         \includegraphics[width=\textwidth]{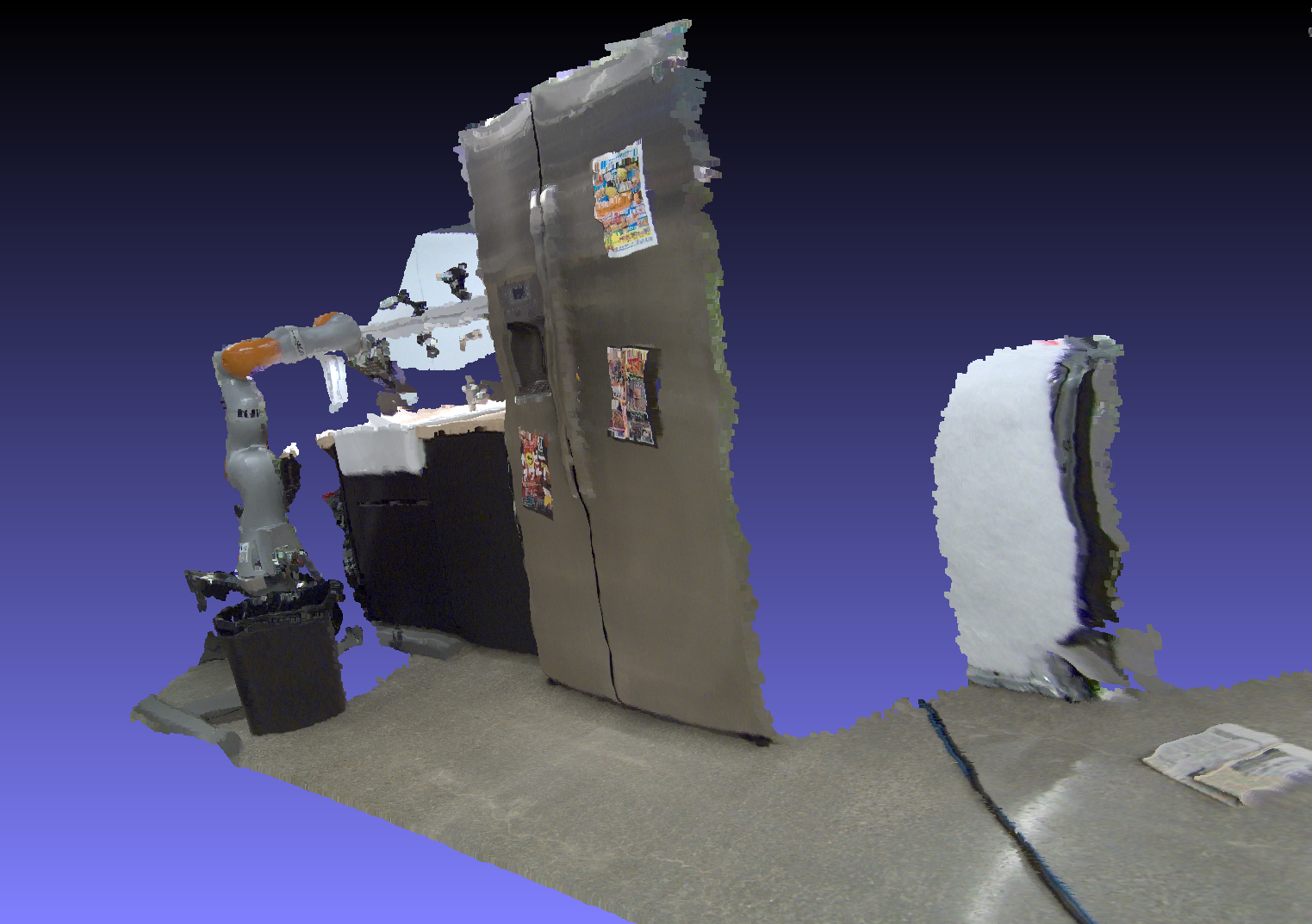}
         \caption{Our point cloud output on image \ref{fig:fridge_rgb} seen from the right.}
         \label{fig:five over x}
     \end{subfigure}
     \hfill
     \begin{subfigure}[b]{0.29\textwidth}
         \centering
         \includegraphics[width=\textwidth]{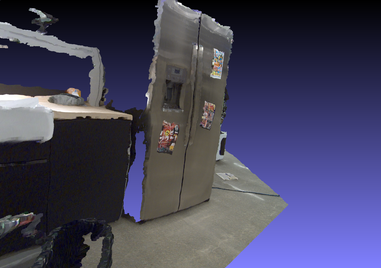}
         \caption{Our point cloud output on image \ref{fig:fridge_rgb} seen from the right.}
         \label{fig:five over x}
     \end{subfigure}
     \hfill
     \begin{subfigure}[b]{0.29\textwidth}
         \centering
         \includegraphics[width=\textwidth]{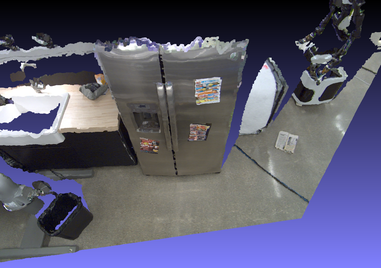}
         \caption{Our point cloud output on image \ref{fig:fridge_rgb} seen from the left.}
         \label{fig:five over x}
     \end{subfigure}
        \caption{Our approach produces substantially more dense depth maps than off-the-shelf sensors, with a larger field of view, while maintaining metric accuracy.}
        \label{fig:single_good_example}
\end{figure*}
\section{Related Work}
\emph{Stereo Matching:}
Stereo depth estimation has a rich history in robotics, the Stanford AI Lab `Cart' being one of the earliest platforms to integrate stereo based depth \cite{hpmoravec_stereocart}. Since then, a large number of advances have been made \cite{scharstein2002taxonomy} to improve speed and quality in a variety of situations. A great deal of algorithmic and implementation effort has gone into making these algorithms run efficiently for online usage \cite{goldberg2002stereo, schauwecker2015sp1}, including optimization to computer architecture as well as the use of FPGAs. However, classical methods face a hard trade-off between efficiency and quality, and the computational cost of achieving results similar to those of more recent learned approaches using engineered methods is prohibitive.

\emph{Learned Depth Estimation}:
Although learning techniques were applied to aid in the process of matching before the widespread use of convolutions (e.g. \cite{kong2004method, zhang2007estimating, spyropoulos2014learning}), these approaches are computationally expensive and produce less accurate results than algorithms leveraging differentiable convolutions in deep models - the earliest of which is \cite{zbontar2016stereo}. This early approach uses semi-global matching and interpolation-based sub-pixel refinement, the shortcomings of which are  well known \cite{scharstein2002taxonomy,stein2006attenuating}. There has been a large effort to develop learned, fully-convolutional models that predict depth from a single view \cite{gordon2019depth,godard2017unsupervised} - monocular, unsupervised models are well known to be far less accurate than stereo-based, supervised models as noted by \cite{godard2017unsupervised,wang2019unos,smolyanskiy2018importance}.

 Fully differentiable models heavily leveraging 2D/3D convolutions achieve the state of the art in real-time performance (see e.g. Middlebury Stereo Benchmark leaderboard \cite{scharstein_szeliski_hirschmuller}). These leverage the high-level structure of classical stereo algorithms \cite{scharstein2002taxonomy}, with all key components replaced by learnable and differentiable alternatives (see Table \ref{table:arch}).
\begin{table}
\begin{tabular}{|p{1.575cm}|p{2.75cm}|p{2.75cm}|}
 \hline
  & \textbf{Classical Stereo} & \textbf{Learned Stereo} \\
  \hline 
  \textbf{Features} & Hand-built e.g. pixel windows, edges, sign of laplacian & Learned, fully convolutional feature extraction \\
  \hline 
  \textbf{Cost-Volume Formation} & Scalar-valued cost function e.g. SAD, SSD & Vector-valued comparison function e.g. elementwise operations \\
  \hline
  \textbf{Cost Aggregation} & Hand-picked 2D/3D filters, functions & Learned 2D/3D convolutions and filters \\
  \hline
  \textbf{Disparity Computation} & Local/Global Optimization & Soft-argmin (see section \ref{subsection:learned_model} for discussion) \\
  \hline
  \textbf{Disparity Refinement} & Hand-engineered filtering e.g. median filtering,occlusion detection, surface-fitting & Fully learnable filters for upsampling and refinement \\
 \hline
\end{tabular}
 \caption{Stereo architecture breakdown: Classical vs Learned}
 \label{table:arch}
\end{table}

 This high level architecture was first presented in \cite{kendall2017end}, and other approaches (\cite{khamis2018stereonet, zhang2018activestereonet, duggal2019deeppruner, chang2018pyramid, smolyanskiy2018importance, cheng2020hierarchical}) build on it with various modifications and variants. Our learned stereo model follows the same high-level structure as these works, but introduces a number of novel modifications and additions that allow us to produce \emph{accurate depth maps at high resolution more efficiently than any prior work} (see Sec. \ref{section:validation} for metric comparisons). Our architecture makes heavy use of components amenable to automatic hardware acceleration and optimization on modern GPUs (similar to \cite{smolyanskiy2018importance}) for maximum efficiency.

\label{section:related_work}
\section{Stereo Matching Algorithm}
\label{section:stereo}
\subsection{Learned Matching Model}
\begin{figure*}[tbh]
    \centering
\begingroup%
  \makeatletter%
  \newcommand*\fsize{\dimexpr\f@size pt\relax}%
  \newcommand*\lineheight[1]{\fontsize{\fsize}{#1\fsize}\selectfont}%
  \setlength{\unitlength}{\textwidth}%
  \global\let\svgwidth\undefined%
  \global\let\svgscale\undefined%
  \makeatother%
  \fontsize{6pt}{7pt}\selectfont
  \begin{picture}(1,0.22171665)(-0.005,0.02)%
    \lineheight{1}%
    \setlength\tabcolsep{0pt}%
    \put(0,0){\includegraphics[width=\unitlength,page=1]{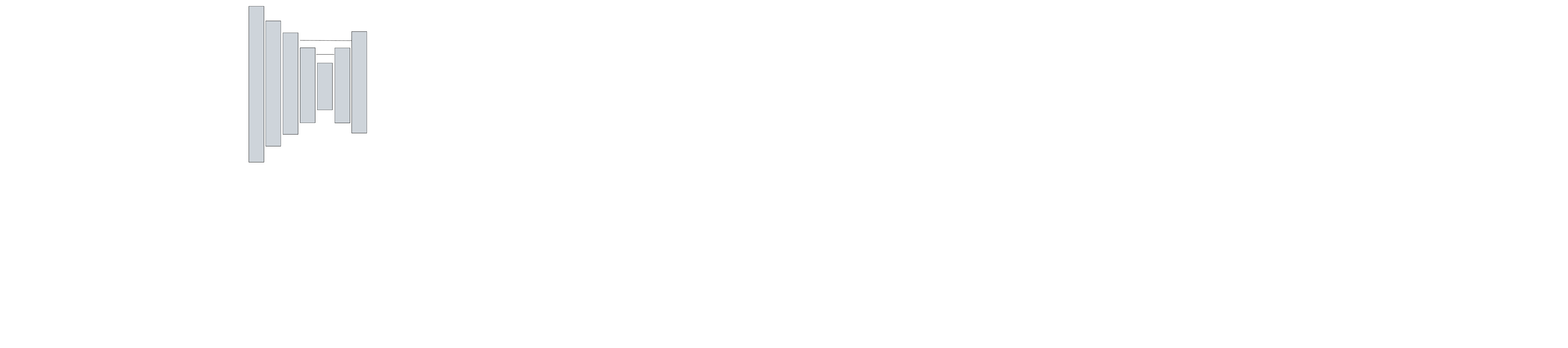}}%
    \put(0.17212712,0.21074799){\color[rgb]{0,0,0}\makebox(0,0)[lt]{\lineheight{1.25}\smash{\begin{tabular}[t]{l}Feature Encoders\end{tabular}}}}%
    \put(0.04216058,0.21164607){\color[rgb]{0,0,0}\makebox(0,0)[lt]{\lineheight{1.25}\smash{\begin{tabular}[t]{l}Left Image\end{tabular}}}}%
    \put(0,0){\includegraphics[width=\unitlength,page=2]{network_diagram_no_text.pdf}}%
    \put(0.03646939,0.0962803){\color[rgb]{0,0,0}\makebox(0,0)[lt]{\lineheight{1.25}\smash{\begin{tabular}[t]{l}Right Image\end{tabular}}}}%
    \put(0,0){\includegraphics[width=\unitlength,page=3]{network_diagram_no_text.pdf}}%
    \put(0.29700608,0.15056264){\color[rgb]{0,0,0}\makebox(0,0)[lt]{\lineheight{1.25}\smash{\begin{tabular}[t]{l}Cost Volume\end{tabular}}}}%
    \put(0,0){\includegraphics[width=\unitlength,page=4]{network_diagram_no_text.pdf}}%
    \put(0.40334992,0.15416875){\color[rgb]{0,0,0}\makebox(0,0)[t]{\lineheight{1.25}\smash{\begin{tabular}[t]{c}3D CV\end{tabular}}}}%
    \put(0.40334992,0.14479597){\color[rgb]{0,0,0}\makebox(0,0)[t]{\lineheight{1.25}\smash{\begin{tabular}[t]{c}Processing\end{tabular}}}}%
    \put(0,0){\includegraphics[width=\unitlength,page=5]{network_diagram_no_text.pdf}}%
    \put(0.52567886,0.09978108){\color[rgb]{0,0,0}\makebox(0,0)[lt]{\lineheight{1.25}\smash{\begin{tabular}[t]{l}Soft Argmin\end{tabular}}}}%
    \put(0.52239339,0.17736575){\color[rgb]{0,0,0}\makebox(0,0)[lt]{\lineheight{1.25}\smash{\begin{tabular}[t]{l}Matchability\end{tabular}}}}%
    \put(0,0){\includegraphics[width=\unitlength,page=6]{network_diagram_no_text.pdf}}%
    \put(0.71567704,0.01817792){\color[rgb]{0,0,0}\makebox(0,0)[t]{\lineheight{1.25}\smash{\begin{tabular}[t]{c}low resolution disparity\end{tabular}}}}%
    \put(0.93725231,0.1165015){\color[rgb]{0,0,0}\makebox(0,0)[t]{\lineheight{1.25}\smash{\begin{tabular}[t]{c}Full resolution disparity\end{tabular}}}}%
    \put(0.7118759,0.14239873){\color[rgb]{0,0,0}\makebox(0,0)[lt]{\lineheight{1.25}\smash{\begin{tabular}[t]{l}Refine Network\end{tabular}}}}%
    \put(0,0){\includegraphics[width=\unitlength,page=7]{network_diagram_no_text.pdf}}%
    \put(0.46311997,0.152116){\color[rgb]{0,0,0}\makebox(0,0)[t]{\lineheight{1.25}\smash{\begin{tabular}[t]{c}2D CV\end{tabular}}}}%
    \put(0.46311997,0.14274323){\color[rgb]{0,0,0}\makebox(0,0)[t]{\lineheight{1.25}\smash{\begin{tabular}[t]{c}Processing\end{tabular}}}}%
    \put(0.22655559,0.1151567){\color[rgb]{0,0,0}\makebox(0,0)[t]{\lineheight{1.25}\smash{\begin{tabular}[t]{c}shared\end{tabular}}}}%
    \put(0.22655559,0.1068392){\color[rgb]{0,0,0}\makebox(0,0)[t]{\lineheight{1.25}\smash{\begin{tabular}[t]{c}weights\end{tabular}}}}%
    \put(0,0){\includegraphics[width=\unitlength,page=8]{network_diagram_no_text.pdf}}%
    \put(0.63470467,0.06660922){\color[rgb]{0,0,0}\makebox(0,0)[t]{\lineheight{1.25}\smash{\begin{tabular}[t]{c}c\end{tabular}}}}%
    \put(0,0){\includegraphics[width=\unitlength,page=9]{network_diagram_no_text.pdf}}%
    \put(0.71000608,0.06686745){\color[rgb]{0,0,0}\makebox(0,0)[t]{\lineheight{1.25}\smash{\begin{tabular}[t]{c}c\end{tabular}}}}%
    \put(0,0){\includegraphics[width=\unitlength,page=10]{network_diagram_no_text.pdf}}%
    \put(0.84027812,0.06614852){\color[rgb]{0,0,0}\makebox(0,0)[t]{\lineheight{1.25}\smash{\begin{tabular}[t]{c}+\end{tabular}}}}%
    \put(0,0){\includegraphics[width=\unitlength,page=11]{network_diagram_no_text.pdf}}%
  \end{picture}%
\endgroup%
    \caption{Network diagram.}
     \label{fig:network_diagram}
\end{figure*}
Our learned model (see Figure \ref{fig:network_diagram}) combines components from prior work and builds on them as follows:
 \begin{enumerate}
     \item \emph{Feature Extraction} 
     The feature extractor is based on a dilated ResNet \cite{koinet}. This gives the model a large receptive field without requiring a layer depth that would prohibit real-time inference at high resolutions. The output of the feature extractor is a 16-dimensional feature map downsampled from the input resolution by a factor of four for the low resolution variant and a factor of eight for high resolution for the high resolution variant.
     \item \emph{Cost Volume Creation} A cross-correlation cost volume \cite{sceneflow} is used to create a 4D feature volume at a configurable number of disparities. 
     \item \emph{Cost Aggregation} The volume is passed through a series of 3D and 2D convolutions to produce a 3D cost volume at the same resolution of the feature extraction stage, similar to \cite{smolyanskiy2018importance}, with fewer operations dedicated to 3D convolution, and more 2D filtering to result in higher quality depth maps at faster rates. 
     \item \emph{Disparity Computation} A differentiable soft argmin operation \cite{kendall2017end} is used to regress a continuous disparity estimate per pixel. A matchability operation \cite{matchability} is used to estimate a confidence of the disparity estimate. 
     \item \emph{Disparity Refinement} A second dilated ResNet \cite{koinet} is used to calculate a disparity residual given the original input image, low resolution disparity, and matchability. The low resolution disparity is bilinearly upsampled to the full input resolution and added to the disparity residual to produce the final disparity estimate.
 \end{enumerate}
More information including a detailed layer description, code implementing this model and videos are available online and as part of supplementary materials. \footnote{Code available at \url{https://sites.google.com/view/stereoformobilemanipulation}.}

\label{subsection:learned_model}

\subsection{Post-Processing}
\label{subsection:filtering}
Output depth pixels are considered valid when they pass a confidence check, and the containing `depth-region' is sufficiently large (2000 px or larger). Confidence is calculated from the output of the matchability module as $\textrm{confidence}= \exp{(\textrm{matchability})}$ \cite{matchability}. The region-size associated with a pixel is computed by running a standard coloring-based algorithm. A thresholding operation with a value of 0.25 for confidence is used to remove disparity estimates with low confidence or small size. 

\section{Sensor-Head Design and Hardware}

For manipulation, we need to be able to grasp objects as small as our gripper can hold, and avoid obstacles conservatively -- to this end we form a voxel map with 2 cm$^2$ voxels, so we require that depth error is at most 1 cm. Furthermore, our robot's maximum reach (slightly further than the average adult human) is about 2 m. 
To achieve this, we selected a stereo camera system based on the Basler acA 2500 \cite{basler} sensors and Sunex DSL318D-650-F2.4 \cite{sunex} lenses, with a 10 cm baseline, giving us a $100 ^\circ$ horizontal $\times$ $80 ^\circ$ vertical rectified image field of view, at 2560 px $\times$ 2048 px resolution.  The cameras are hardware triggered to produce synchronized capture.

\begin{figure}
    \includegraphics[width=0.475\textwidth]{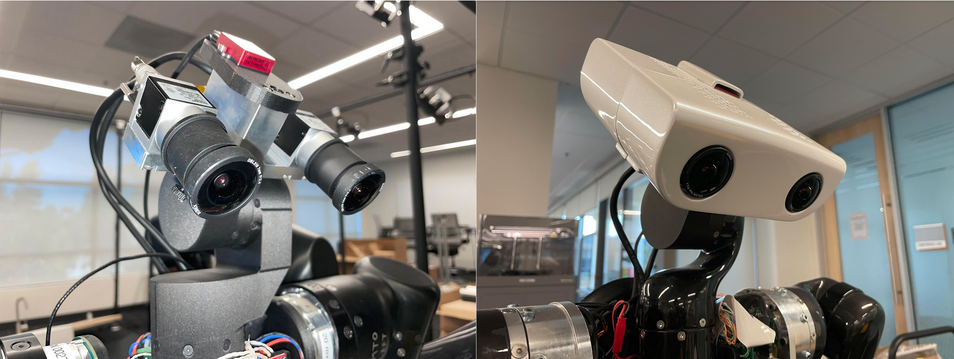}
    \label{fig:stereo_head}
    \caption{Stereo camera pair used on our robot, side by side with and without covering.
    }
\end{figure}
\label{section:hardware}
\section{Data Collection}
\begin{figure}[htb]
     \begin{subfigure}[b]{0.25\textwidth}
         \includegraphics[width=0.9\linewidth]{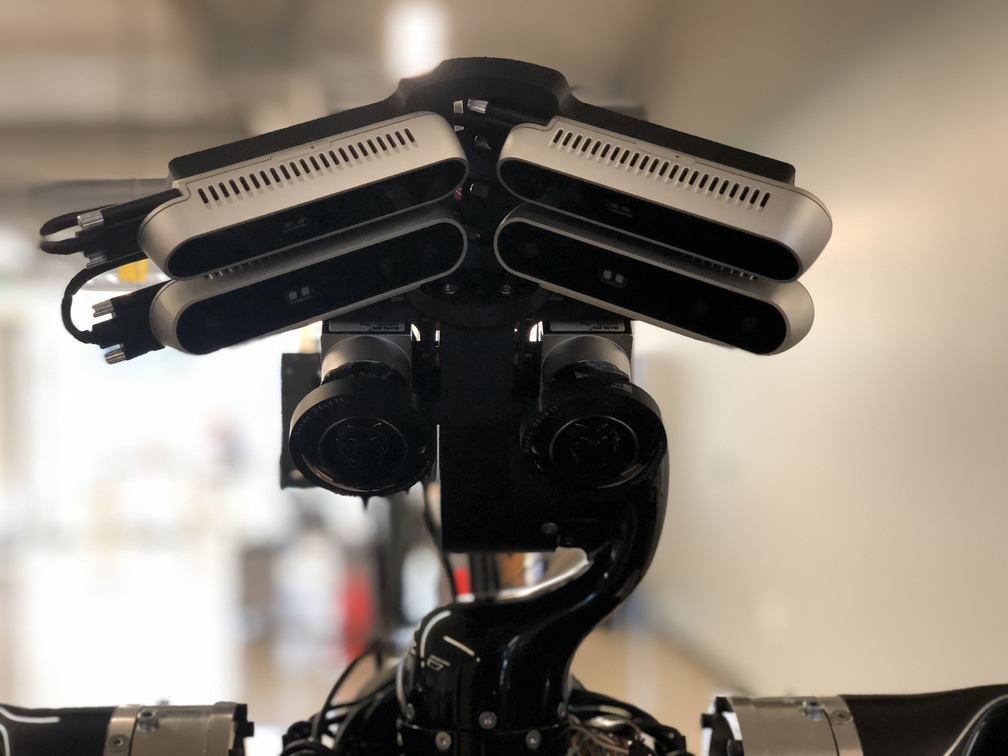}
         \caption{Data collection head v0}
         \label{fig:collection_head_v0}
     \end{subfigure}%
     \begin{subfigure}[b]{0.25\textwidth}
         \includegraphics[width=0.9\linewidth]{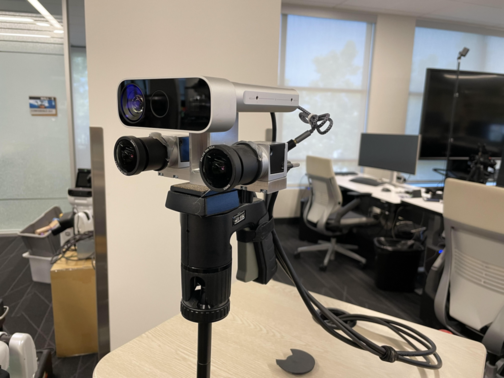}
         \caption{Data collection head v1}
         \label{fig:collection_head_v1}
     \end{subfigure}
        \caption{The various iterations on the data collection heads used for collection ground truth stereo data. \ref{fig:collection_head_v0} was the first iteration that used 4 Intel Realsense D415 cameras to cover the wide field of view of the two hi-res Basler color cameras. \ref{fig:collection_head_v1} is the latest iteration that replaced the 4 Intel Realsense cameras with a single Microsoft Azure Kinect sensor that offers roughly the same FOV coverage of the Basler color cameras.}
        \label{fig:data_collection_heads}
\end{figure}

To train our stereo network, a significant amount of real data was collected with accurate depth labels at each (or most) pixels. To achieve this, we constructed a custom "data collection head" that consisted of the two Basler color cameras coupled with another external sensor to serve as the ground truth label. The first iteration of the data collection head is shown in figure \ref{fig:collection_head_v0}, where 4 Intel Realsense D415 stereo camera units were configured to optimally cover (roughly) the same field of view as the Basler stereo cameras. Depth from the Intel Realsense cameras is reprojected into the Basler cameras to get a ground truth label at the reprojected pixel location. Figure \ref{fig:fridge_realsense} illustrates the coverage of the reprojected depth from the Realsense cameras. 

The second version of the collection head we developed utilized a Microsoft Azure Kinect sensor which replaced the Intel Realsense D415 array we had previously used. This sensor provided roughly the same coverage as the Basler cameras with the added benefit of being able to synchronize capture with an external trigger source. Figure \ref{fig:collection_head_v1} shows the latest iteration of the data collection head. With the synchronous capture problem now resolved, this unlocked a much faster way of collecting data by capturing continuously. To date, we've been able to scan 10 homes, resulting in 410 unique scans of 154 scenes. Of those scans, we've extracted 323,192 useful stereo frames labeled with ground-truth depth from the Kinect sensor. A subset of these are carefully selected for use in training (see Section \ref{section:training}).
\label{subsection:data_collection}
\section{Training}
\label{section:training}
\subsection{Mixing Real and Synthetic Data}

\begin{table}
  \centering
  \begin{tabular}{|c|c|c|}
 \hline
 Data Type  & Middlebury EPE & Our Real Data EPE\\
 \hline
 Real & 4.65 & 0.75 \\
 \hline
 Synthetic & 3.04 & 1.47 \\
 \hline
 Mixed & 3.05 & 0.83 \\
 \hline
\end{tabular}
  \caption{End-point-disparity-error (EPE) on Middlebury benchmark and on our data, as a function of training data distribution.}
  \label{table:metric_accuracy}
\end{table}

We found that using a diverse combination of synthetic and real data gave the best benchmark and real-world performance on our robot (See Table \ref{table:metric_accuracy} for comparison, the specific contents of the Mixed and Synthetic datasets are detailed in Section \ref{subsection:training_procedure}). Synthetic data was useful because of the large variety of scenes, objects, materials, and lighting. Using simulation we are able to get perfect ground truth data on non-lambertian surfaces such as glass and metal that was missing in the real dataset.

The Randomixed Texture (RT) synthetic dataset contains procedurally generated indoor manipulation scenes containing a room, the robot, furniture and objects sampled from the ShapeNet database. We rendered the scenes using a simple and fast OpenGL renderer with random image based textures and procedural material properties.

For the Synthetic Flying Things (SFT) synthetic dataset we found it useful to use a higher fidelity renderer to capture the complex lighting effects on shiny materials. Blender's Eevee rendering backend was a good sweet spot between fidelity and speed. We used randomized scene placement, materials, and lighting with a bias towards reflective materials for this dataset.

The Facebook Replica Synthetic Dataset \cite{replica19arxiv} is a set high quality reconstructions of a variety of indoor spaces. Each reconstruction has clean dense geometry, high resolution and high dynamic range textures, glass and mirror surface information. We sample views randomly from all of the available scenes to generate a diverse set of realistic samples.

We also use a commercial synthetic dataset containing many indoor scenes with semantically plausible scene placement, materials, and lighting. Due to the semantically consistency the apparent variety is lower, but the images qualitatively look normal compared to flying-things style datasets. They use a high quality path tracing back-end which produces high quality global illumination effects. A large number of companies generate data of this kind \footnote{E.g. \url{http://www.coohom.com}, \url{http://www.datagen.tech}}.  

\begin{figure}
\centering
\begin{tabular}{cc} 
\includegraphics[scale=0.35]{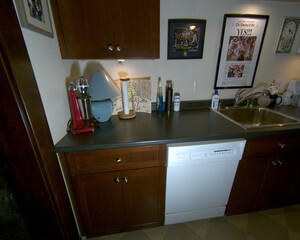} &
\includegraphics[scale=0.35]{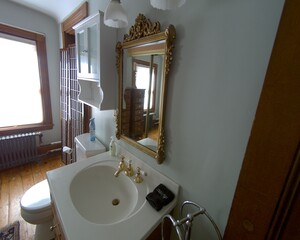} \\
\includegraphics[scale=0.35]{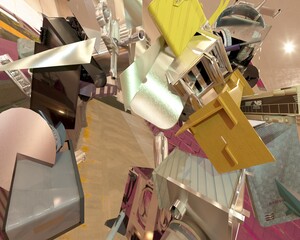} &
\includegraphics[scale=0.35]{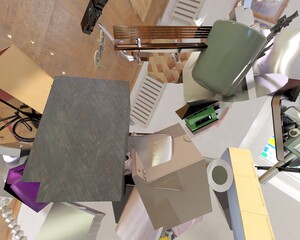} \\
\includegraphics[scale=0.35]{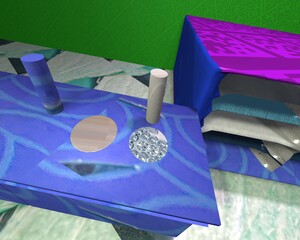} &
\includegraphics[scale=0.35]{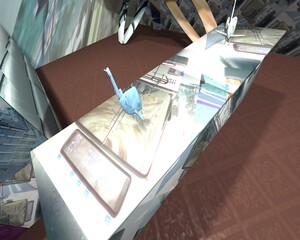} \\
\includegraphics[scale=0.35]{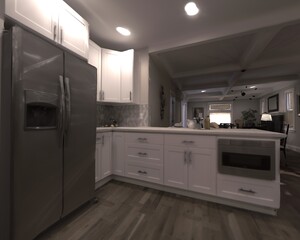} &
\includegraphics[scale=0.35]{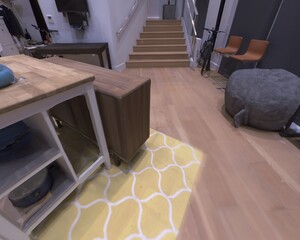} \\
\end{tabular}
\caption{Examples of real (top), shiny-things (second row), random-thing (third row), and Replica training data (bottom)}
\end{figure}
\label{subsection:mix}

\subsection{Loss Functions}
A smooth L1 loss function is used on the soft argmin and disparity refinement \cite{chang2018pyramid}. To account for the effect of the different disparity scales at each resolution and the variability of the disparity range in each batch element, the loss is scaled by the mean and standard deviation per batch i.e. 
$$L_{\textrm{disparity}}(d, \hat{d}) =  \sum_{b=1}^{\textrm{batchsize}}\frac{\left(\sum_{i=1}^{N_\textrm{pixels}}\textrm{smooth} _{L_1}(d_i - \hat{d_i})\right)}{\mu_b + 2\sigma_b}$$ where $d_i$ is the ground truth disparity at pixel index $i$ and $\hat{d}_i$ is the corresponding prediction, and $\mu_b,\: \sigma_b$ are the mean and standard deviations of the ground truth disparities within batch $b$, and $N_\textrm{pixels}$ is the number of pixels in an image.
A noise sampling cross entropy loss ($L_{nsce}$)\cite{chen2020noise} is used to regularize the cost volume and promote unimodal peaks for the soft argmin operation. A smoothness loss $L_\textrm{smooth}$\cite{smolyanskiy2018importance} is used to promote smoothness of the output in low texture areas of the image.
The overall loss is

\begin{align*}
    \textrm{Loss} = &\lambda_1  L_{\textrm{disparity}}(d_{hr}, \hat{d}_{hr}) + \lambda_2L_{\textrm{disparity}}(d_{lr}, \hat{d}_{lr})\\
    &+ \lambda_3  L_{nsce} + \lambda_4 L_\textrm{smooth}.
\end{align*}
Where $d_{hr},\: d_{lr}$ are the high-resolution and low-resolution disparities respectiviely. For all experiments, $\lambda_1 = 100.0, \lambda_2 = 100.0, \lambda_3 = 0.2, \lambda_4 = 20.0$.
\subsection{Training Procedure}
\label{subsection:training_procedure}
All training was done using the Adam optimizer ($\beta_1$ = 0.9, $\beta_2$ = 0.999). An initial learning rate of 0.001 is used for pre-training. For fine tuning, an initial learning rate of 0.0001 is used. A polynomial learning rate decay with a factor of 0.9 is used.

For all datasets, flip and color jitter data augmentation are used. Noise, blur, chromatic aberration, and bayer/debayer augmentation are used in addition for synthetic datasets.

For deployment on the mobile manipulation robot, a maximum disparity of 384 and a cost volume downsample of eight was used. The network was pre-trained for 100 epochs with a mix of 10000 samples from the Replica synthetic dataset, 20000 samples from the RT synthetic dataset, and 20000 samples from the SFT synthetic dataset. The network was fine-tuned for 100 epochs with a mix of 5000 samples from the Replica synthetic dataset, 10000 samples from the RT synthetic dataset, 10000 samples from the SFT synthetic dataset, 20000 samples from a commercial synthetic dataset, and 6000 samples from the real camera data collection system. A crop size of 1440x896 at a batch size of 16 was used for both pre-training and fine-tuning.

\section{Experimental Validation}
\label{section:validation}

\subsection{Metric Accuracy}
We constructed a simple experiment to spot check the range accuracy of our network's reported depth. We placed our stereo camera pair statically facing a flat wall with an April tag taped to it to provide some added texture. A single beam, laser range finder was used to measure the distance to the wall, which was then treated as ground truth. A range reading from the stereo network was taken at the approximate pixel location of the visible red laser beam and used to record $E_1$, the error between the measured depth and the laser ground truth. A user additionally solved for the sub-pixel correspondences between left and right images of the same point and provided a hand-measured estimate of the stereo disparity and depth to record $E_2$, the error between the user estimated depth and the laser ground truth. This process was repeated for a number of different depths. 

\label{subsection:metric_accuracy}
\subsection{Task Performance}
\label{subsection:task}
\begin{figure*}[htb]
     \centering
     \begin{subfigure}[b]{0.3\textwidth}
         \centering
         \includegraphics[width=\textwidth]{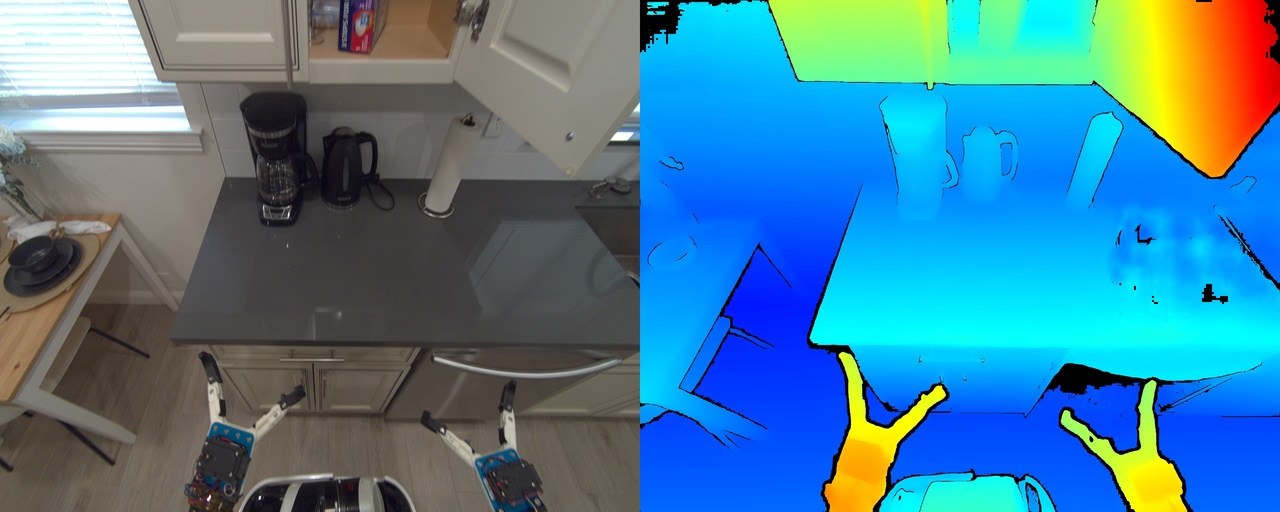}
         \caption{Wiping task: depth on reflective surfaces like countertop and coffee pot.}
         \label{fig:task_coffee_pot}
     \end{subfigure}
     \hfill
     \begin{subfigure}[b]{0.3\textwidth}
         \centering
         \includegraphics[width=\textwidth]{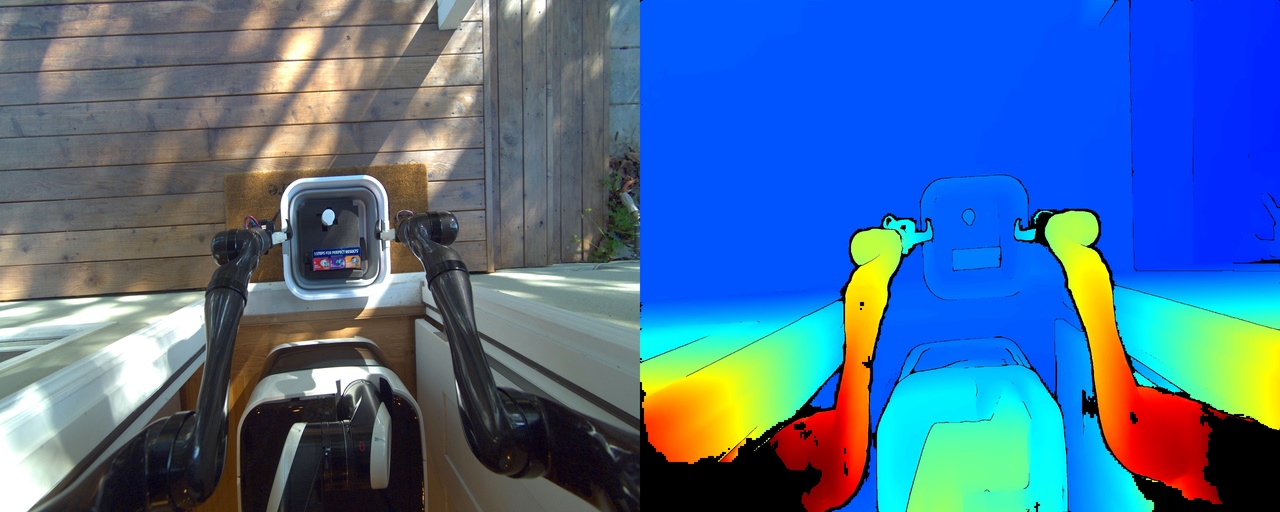}
         \caption{Grocery task: depth in outdoor lighting environments.}
         \label{fig:task_grocery}
     \end{subfigure}
     \hfill
     \begin{subfigure}[b]{0.3\textwidth}
         \centering
         \includegraphics[width=\textwidth]{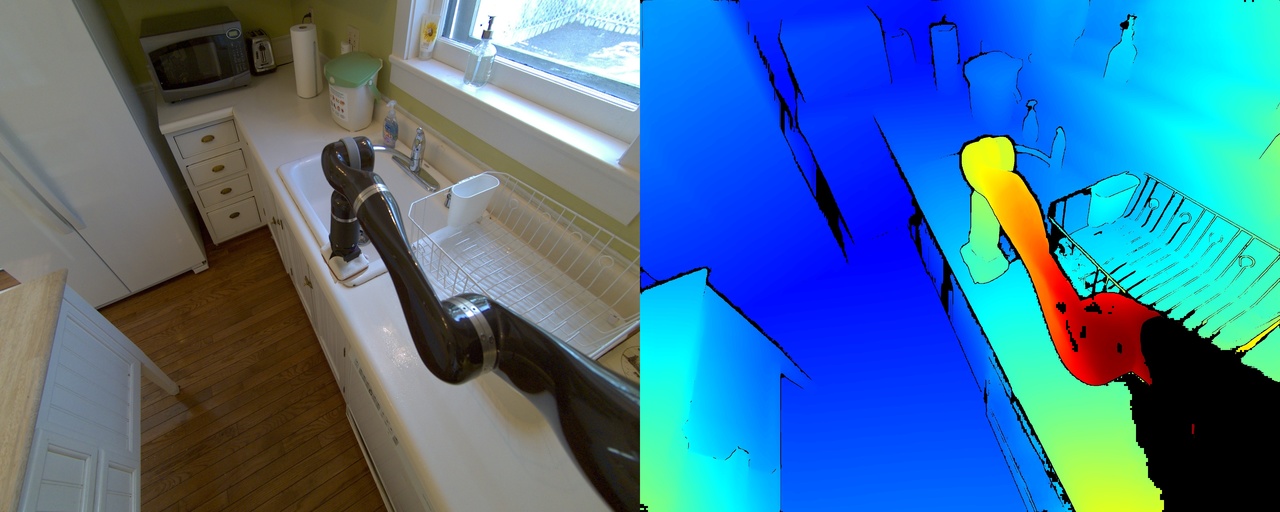}
         \caption{Wiping task: depth on countertop, thin-wired dish rack, and clear bottles.}
         \label{fig:task_kitchen_counter}
     \end{subfigure}
     \begin{subfigure}[b]{0.3\textwidth}
         \centering
         \includegraphics[width=\textwidth]{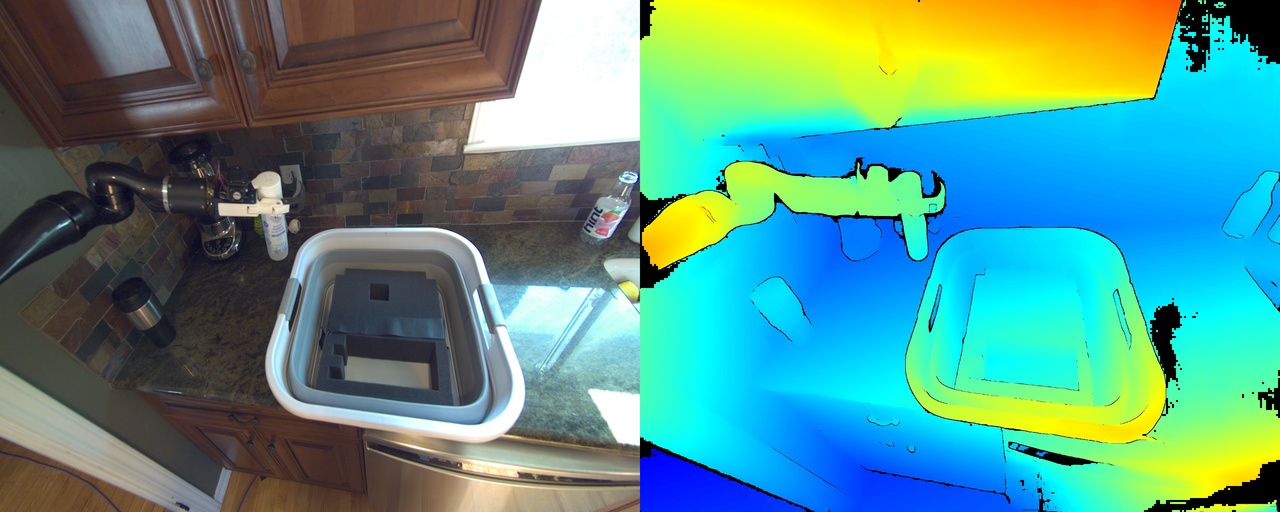}
         \caption{Grocery task: depth on reflective arm in the presence of extreme light exposure.}
         \label{fig:task_shampoo}
     \end{subfigure}
     \hfill
     \begin{subfigure}[b]{0.3\textwidth}
         \centering
         \includegraphics[width=\textwidth]{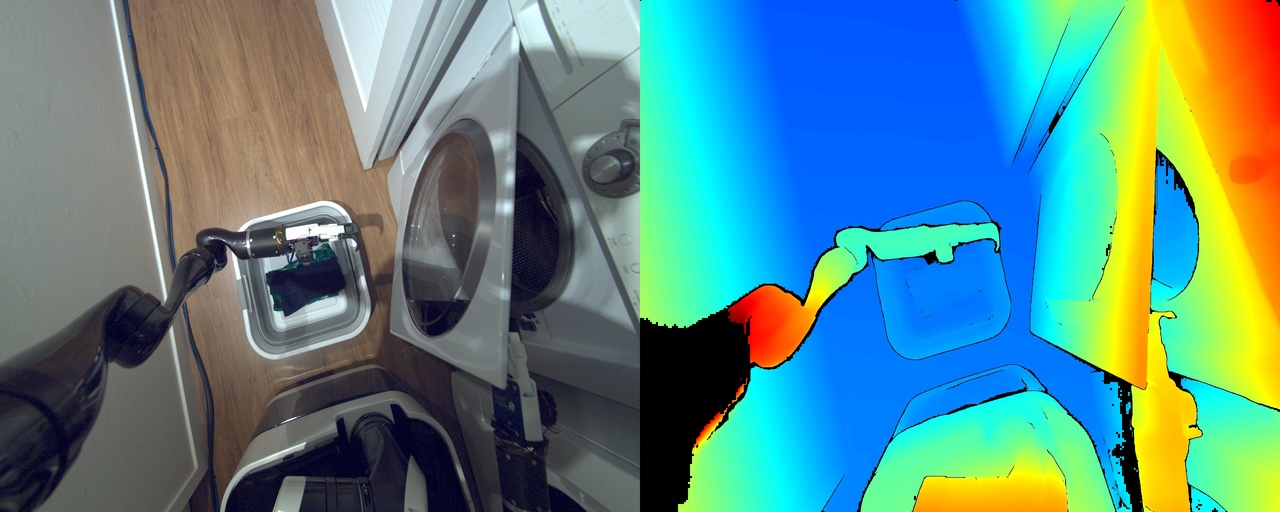}
         \caption{Laundry task: depth on reflective and curved surfaces of the washer and dryer.}
         \label{fig:task_laundry}
     \end{subfigure} 
     \hfill
     \begin{subfigure}[b]{0.3\textwidth}
         \centering
         \includegraphics[width=\textwidth]{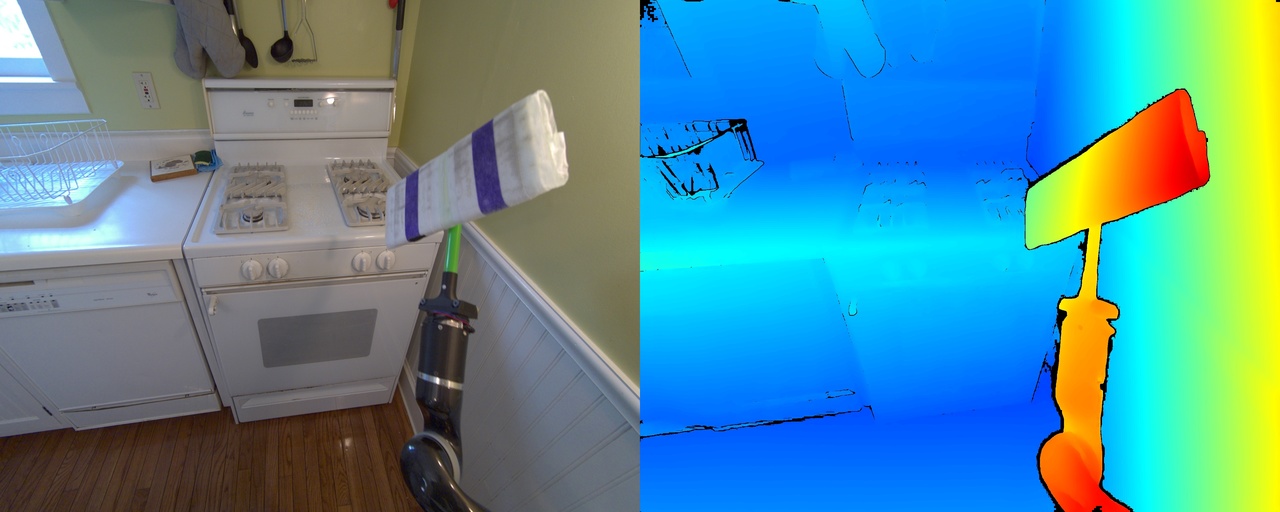}
         \caption{Wiping task: depth on reflective stove top and near field arm of the robot.}
         \label{fig:five over x}
     \end{subfigure}
        \caption{Various manipulation tasks performed by our general purpose robot using the described learned stereo network.}
        \label{fig:tasks_learned_stereo}
\end{figure*}
Our described stereo network has been integrated into our general purpose robot system \cite{bajracharya2020mobile} as the primary exteroceptive perception sensor for performing household tasks. Using only our dense stereo depth for localization, mapping, manipulation, and motion planning, our robot is able to complete a number of various household tasks. Figure \ref{fig:tasks_learned_stereo} illustrates a few images taken from a number of field tests performed in real homes. Such tasks include a "grocery task" where delivered items are taken from the doorstep and organized within the home, a "wiping task" where surfaces in the kitchen are cleaned with a cleaning tool, and a "laundry task" where a dryer must be first opened then items are taken out of the dryer and placed in a laundry bin.

\subsection{Speed}
Robotic manipulation systems require high camera framerates (5 Hz or higher) for effective obstacle avoidance and closed-loop control -- moreover, given the large number of different computational demands on limited on-robot compute, the processing budget per frame for stereo-depth must be minimized. Our learned model is optimized to produce high quality depth maps as efficiently as possible, leveraging operations and sequences of operations amenable to GPU execution and optimized inference using TensorRT (TRT) \footnote{\url{https://developer.nvidia.com/tensorrt}}. While our inference is fast using PyTorch \cite{pytorch} native runtime, which runs all 32-bit floating point operations using CUDA where possible C++ otherwise, it is exceedingly performant when run using optimized inference. Table \ref{table:speed} shows the speed of our model for various input resolutions, corresponding to different data sources. The runtimes show the wall-time taken to produce the raw output of the learned model, and don't include any post-processing. The number of disparities (ndisp) used varies with the maximum disparity in each dataset's ground-truth. The speeds are shown for 3 different commonly used GPU (Titan Xp, Titan RTX, and Tesla V100), with varying levels of optimization (native PyTorch, TRT optimized with 32-bit operations, and TRT optimized with 16-bit operations).

 To our knowledge, there is no prior work that has produced such high-quality, high-resolution depth maps at these rates. Though other work outperforms us in terms of metric quality, we are at least 15X faster than any approach producing depth maps of similar or better quality at high resolutions (see Section \ref{subsection:comparison}).
\begin{table*}
\centering
  \begin{tabular}{ |p{1.35cm}|p{0.35cm}p{0.35cm}p{0.35cm}p{.5cm}|p{.75cm}p{.75cm}p{.75cm}|p{.75cm}p{.75cm}p{.75cm}|p{.75cm}p{.75cm}p{.75cm}|}
  \hline
 \multicolumn{5}{|c|}{} & \multicolumn{3}{|c|}{Titan Xp} & \multicolumn{3}{|c|}{Titan RTX}& \multicolumn{3}{|c|}{Tesla V100}\\
 \hline
 Image Source & rows & cols & ndisp & CV Scale & PyTorch & TRT32 & TRT16 & PyTorch & TRT32 & TRT16 & PyTorch & TRT32 & TRT16  \\
 \hline \rule{0pt}{0.75\normalbaselineskip}
 Ours  & 2048 & 2560 & 384 & 8 & 232 & 157 & 164 & 128 & 92 & 33 & 113 & 77 & 28 \\
 \hline\rule{0pt}{0.75\normalbaselineskip}
 Kitti & 376 & 1242 & 256 & 4& 57 & 43 & 39 & 38 & 24 & 8 & 39 & 22 & 8\\
 \hline\rule{0pt}{0.75\normalbaselineskip}
 Sceneflow  & 540&	960&	256	&4&	60	& 41 &	43 &	41&	28 & 9		& 41	& 25 & 9 \\
 \hline
Middlebury&	1984&	2872&	512	&8&	268& 194&196&	158&	118&	41&	142&	102&37\\
 \hline
\end{tabular}
  \caption{Learned Model GPU Inference Runtimes: Different Data Sources on Different GPUs using Different Frameworks.}
  \label{table:speed}
\end{table*}

\subsection{Comparison on Stereo Benchmarks}
\label{subsection:comparison}
In order to compare our algorithm against prior work, we've trained and evaluated\footnote{Note that we make all of these comparisons only using the raw output of our learned model - the metrics and runtimes presented do not take any parallel, on-robot post-processing (see Section \ref{subsection:filtering}) into account} it on publicly available benchmarks including the Sceneflow `Flying Things' dataset\cite{sceneflow}, and the Middlebury Dataset \cite{middlebury} . These comparisons show that our performance in terms of metrics is comparable (and in some cases better than) prior work, while our efficiency is dramatically higher than the works we build on. For each benchmark, we train a model with a number of disparities, and cost-volume downsample rate suited to the dataset. Our relative metric performance is best on the Middlebury dataset - these scenes and images are most similar to those of our camera and our use-case. Our model is the fastest across the board - while there are other approaches that are significantly slower than ours that perform better, to our knowledge there are no methods that are similarly efficient at high resolution with better metric performance. All other prior work is far too slow to be used as part of a robotic system -- while other works produce depth maps at 5 MP at 2Hz or slower, we are able to produce depth maps at \emph{15 Hz} at this resolution.  

\subsubsection{Sceneflow}
The sceneflow \cite{sceneflow} FlyingThings3D stereo dataset consists of a large number of ShapeNet objects placed in random poses with random photos used as background texture. For this experiment, a maximum disparity of 256 and a cost volume downsample of four was used. The network was trained for 100 epochs with a crop size of 896x480 at a batch size of 24. Though this data contains a great deal of low (no) texture objects, and its images are very low-resolution, our performance improves upon the results from \cite{chang2018pyramid, kendall2017end} and is close to some of the best performing models. Our approach also scales far more efficiently with resolution.
\begin{table}
  \centering
  \begin{tabular}{ |c|c|c|c|}
 \hline
 Method & Runtime (ms) & EPE px & $\%$ Bad (1.0) \\
 \hline
 PSMNet \cite{chang2018pyramid}  &   400  & 1.09 & 12.1 \\
 GCNet \cite{kendall2017end} & 900 & 1.84 & 15.6\\
 LEASNet \cite{cheng2020hierarchical} & 300 & 0.78 & 7.82\\
 AANet \cite{xu2020aanet} & 70 & 0.87 & 9.3 \\
 \hline 
 Ours & \textbf{40} & 0.936 & 10.0 \\
 \hline
\end{tabular}
  \caption{Performance on Sceneflow FlyingThings \cite{sceneflow} test set.}
  \label{table:flyingthings}
\end{table}

\subsubsection{Middlebury}
The middlebury dataset consists of high-resolution, densely labeled indoor scenes of high complexity \cite{middlebury}. This benchmark is the best suited to our work, as the resolution and distribution of imagery most closely matches our target use cases. For this experiment, a maximum disparity of 512 and a cost volume downsample-rate of eight was used. The network was pretrained for 1000 epochs with a mix of 50000 samples of custom synthetic data (see Section \ref{subsection:mix}). The network was fine-tuned for 200 epochs with a mix of 25000 samples of custom synthetic data, 20000 samples of commercial synthetic data, 6000 samples from the real camera data collection system, and the Middlebury training set repeated 100 times per epoch. A crop size of 1440x896 at a batch size of 16 was used for both pre-training and fine-tuning. The runtimes shown are those reported on the benchmark website \footnote{\url{https://vision.middlebury.edu/stereo/eval3}} - we are significantly more efficient than the prior work we build upon. With TensorRT optimization, our model runs \emph{more than 15 times faster than the fastest submission to do better than ours} -- in other words, at 5MP resolution, the best performing fast models can run at about 2HZ, while our model runs at \emph{30 HZ}.

\begin{table*}
\centering
\begin{tabular}{ |c|ccccccccccccc|}
 \hline
 \multirow{2}{*}{Method} & \multicolumn{2}{c}{Bad 2.0 ($\%$)}  &\multicolumn{2}{c}{Bad 4.0 ($\%$)} &\multicolumn{2}{c}{Avg Err (px)} & \multicolumn{2}{c}{RMSE (px)}& \multicolumn{2}{c}{A90 (px)} &
 \multicolumn{2}{c|}{A95 (px)} & \multirow{2}{*}{Runtime (ms/mp)}  \\
   & Nocc & All & Nocc & All & Nocc & All & Nocc & All & Nocc & All & Nocc & All & \\
 \hline \rule{0pt}{0.75\normalbaselineskip}
  PSMNet \cite{chang2018pyramid} & 42.1 & 47.2 & 23.5 & 29.2 & 6.68 & 8.78 & 19.4 & 23.3 & 17.0 & 22.8 & 31.3 & 43.4 & 2620 \\
 iResNet \cite{iResNet} & 22.9 & 29.5 & 12.6 & 185 & 3.31 &4.67 & 11.3 & 13.9 & 6.61 & 10.6 & 12.5 &20.7 & 270 \\
 AANet+ \cite{xu2020aanet} & 15.4 & 22.0 & 10.8 & 16.4 & 6.37 & 9.77 & 23.5 & 29.4 & 7.55 & 29.3 & 48.8 & 76.1 & 2480\\
 LEAStereo \cite{cheng2020hierarchical}& 7.15 & 12.1 & 2.75 & 6.33 & 1.43 &2.89 &8.11 &13.7 &1.68 &2.62 &2.65&6.35 & 2530 \\ 
 DeepPruner \cite{duggal2019deeppruner}& 30.1 & 36.4  &15.9 &21.9 &4.8 &6.56 & 14.7 & 18.0 &10.4 &17.9 &23.6&33.1 & 410\\
 HITNet \cite{hitnet}& 6.06 & 12.8& 6.46&12.8&1.71&3.29&9.97&14.5&2.32&3.92&4.26&11.4&110\\
 HSMNet \cite{hsmnet}& 10.2 & 16.5&10.2&16.5&2.07&3.44&10.3&13.4&2.12&4.26&4.32&17.6&100\\
 AdaStereo \cite{adastereo}&13.7&19.8&13.7&19.8&2.22&3.39&10.2&13.9&2.61&4.69&5.67&13.7&130\\
 \hline
 \multirow{2}{*}{Ours} & \multirow{2}{*}{12.7} & \multirow{2}{*}{17.4} & \multirow{2}{*}{7.26} &\multirow{2}{*}{11.0}& \multirow{2}{*}{3.6}&\multirow{2}{*}{5.77}&\multirow{2}{*}{14.2}&\multirow{2}{*}{19.7}&\multirow{2}{*}{3.82}&\multirow{2}{*}{14.5} &\multirow{2}{*}{21.0} &\multirow{2}{*}{35.5} & TRT16: \emph{\textbf{6.5}}\\ & & & & & & & & & & & & & PyTorch: 30\\ 
 \hline
\end{tabular}
 \caption{Performance on Middlebury 2014 Stereo \cite{middlebury} test set.}
 \label{table:middlebury}
\end{table*}
\subsection{Failure Cases}
\begin{figure}
\centering
\begin{tabular}{cc} 
\includegraphics[width=0.45\linewidth]{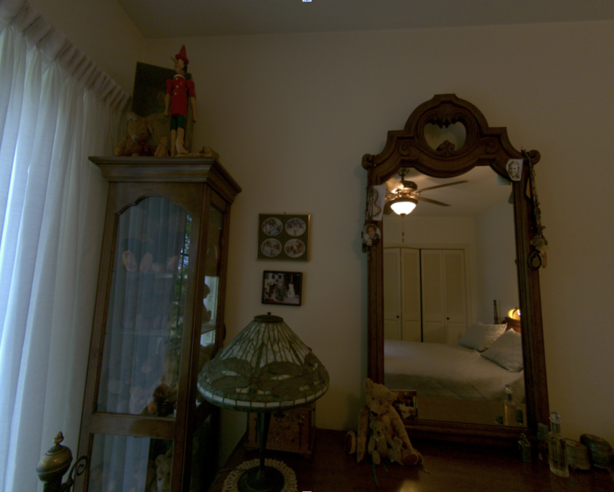} &
\includegraphics[width=0.45\linewidth]{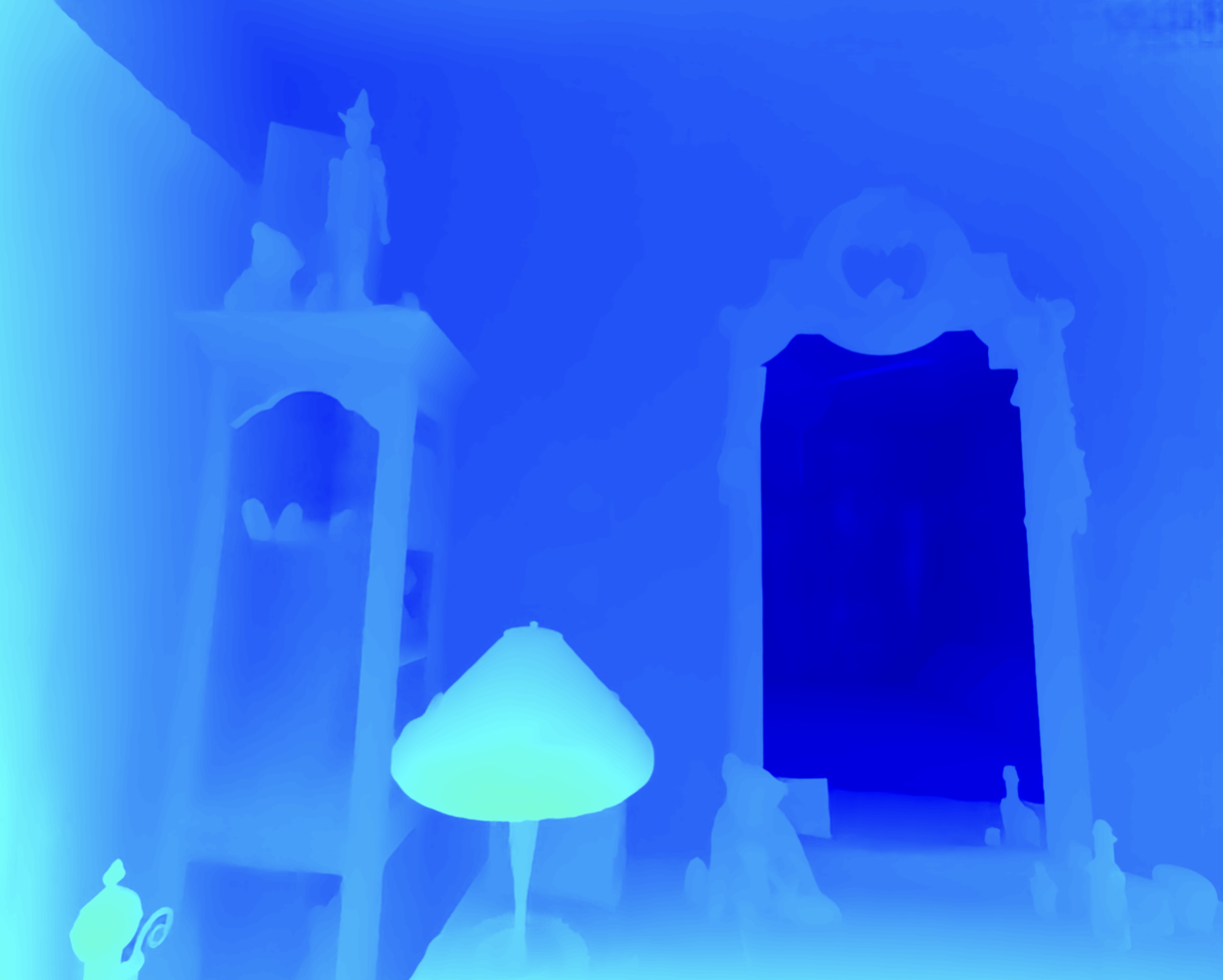} \\
\includegraphics[width=0.45\linewidth]{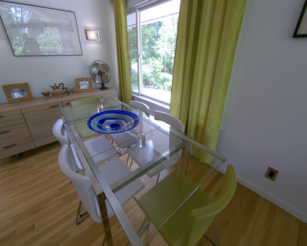} &
\includegraphics[width=0.45\linewidth]{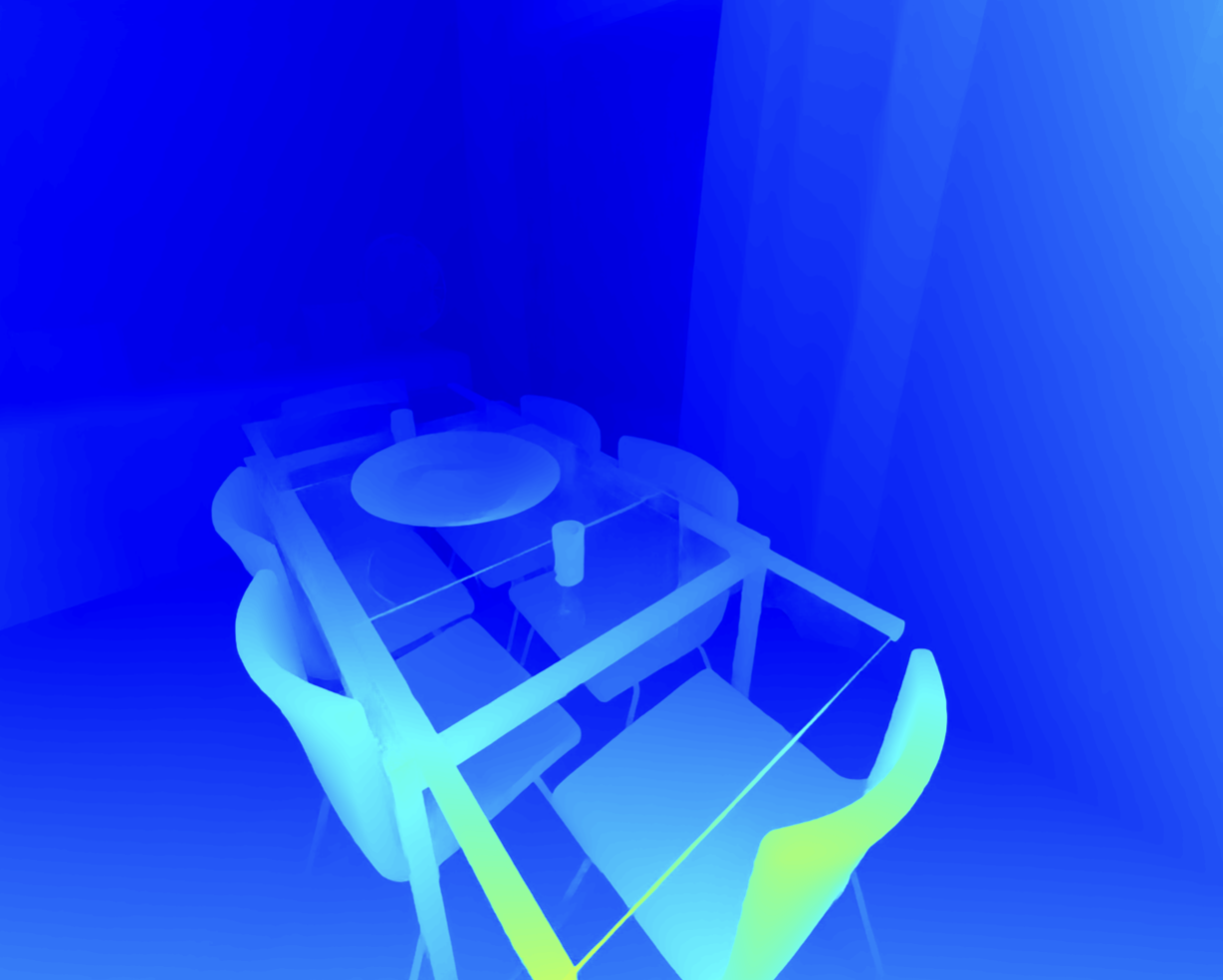} \\
\includegraphics[width=0.45\linewidth]{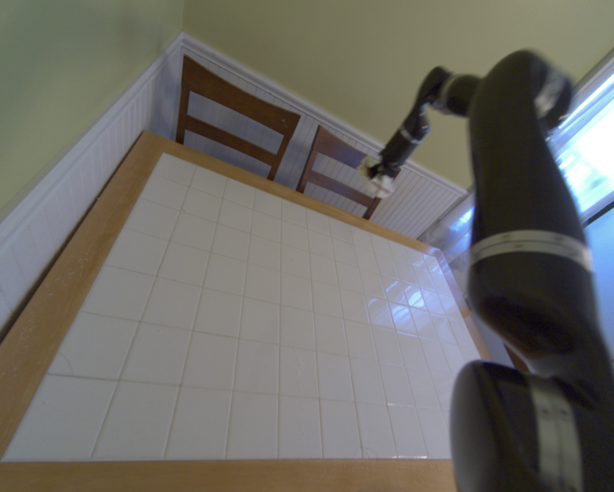} &
\includegraphics[width=0.45\linewidth]{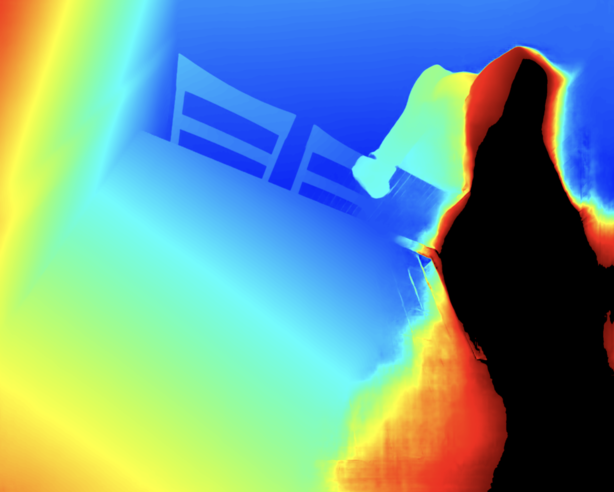}\
\end{tabular}
\caption{Failure cases for our algorithm. Top: Large mirror with depth predicted behind the camera. Middle: A clear glass table that we see straight through. Bottom: The arm and shoulder are so close to the sensor that stereo matching becomes very challenging.}
\label{fig:failures}
\end{figure}

While our algorithm provides substantially improved depth maps compared to existing approaches and depth sensors, it still struggles with some scenes and surfaces. Examples of these are shown in Figure \ref{fig:failures}, and include large mirrors, large glass clear surfaces and objects very close (requiring more than 400 disparities) to the sensor. While we continue to pursue these failure cases, they present fundamental challenges. Improving our performance in these cases remains an area of research.

\section{Conclusion}
We presented a system for generating stereo depth maps that are amenable to use for robotic manipulation in unstructured environments. Our learned model is capable of producing depth maps of high accuracy much faster than comparable approaches, heavily leveraging GPU parallelism and optimization. In order to produce accurate depth on the complex variety of surfaces and objects we encounter, we identified that a mix of large quantities of synthetic and real labeled data was required. We performed a careful search based on first principles to identify a sensor-head that satisfies our field-of-view requirements and makes stereo-depth estimation straight-forward. We showed that our approach performs well by describing it's use as part of robotic manipulation tasks, measuring its metric accuracy in the real world, and showing how we compare to state-of-the-art methods on a variety of stereo-benchmark datasets.

In future work, we intend to investigate integrating information across time and space to produce even more accurate depth maps over larger ranges. 
\label{section:conclusion}
\bibliographystyle{IEEEtran}                          
\bibliography{IEEEabrv,references}  

\begin{thebibliography}{10}
\providecommand{\url}[1]{#1}
\csname url@rmstyle\endcsname
\providecommand{\newblock}{\relax}
\providecommand{\bibinfo}[2]{#2}
\providecommand\BIBentrySTDinterwordspacing{\spaceskip=0pt\relax}
\providecommand\BIBentryALTinterwordstretchfactor{4}
\providecommand\BIBentryALTinterwordspacing{\spaceskip=\fontdimen2\font plus
\BIBentryALTinterwordstretchfactor\fontdimen3\font minus
  \fontdimen4\font\relax}
\providecommand\BIBforeignlanguage[2]{{%
\expandafter\ifx\csname l@#1\endcsname\relax
\typeout{** WARNING: IEEEtran.bst: No hyphenation pattern has been}%
\typeout{** loaded for the language `#1'. Using the pattern for}%
\typeout{** the default language instead.}%
\else
\language=\csname l@#1\endcsname
\fi
#2}}

\bibitem{velodyne_lidar_2018}
\BIBentryALTinterwordspacing
\emph{Velodyne Lidar}, Velodyne Lidar, 2018, rev S3. [Online]. Available:
  \url{https://velodynelidar.com/downloads/}
\BIBentrySTDinterwordspacing

\bibitem{schauwecker2015sp1}
K.~Schauwecker, ``Sp1: Stereo vision in real time.'' in \emph{MuSRobS@ IROS},
  2015, pp. 40--41.

\bibitem{scharstein2002taxonomy}
D.~Scharstein and R.~Szeliski, ``A taxonomy and evaluation of dense two-frame
  stereo correspondence algorithms,'' \emph{International journal of computer
  vision}, vol.~47, no.~1, pp. 7--42, 2002.

\bibitem{bamji2018impixel}
C.~S. Bamji, S.~Mehta, B.~Thompson, T.~Elkhatib, S.~Wurster, O.~Akkaya,
  A.~Payne, J.~Godbaz, M.~Fenton, V.~Rajasekaran, \emph{et~al.}, ``Impixel 65nm
  bsi 320mhz demodulated tof image sensor with 3$\mu$m global shutter pixels
  and analog binning,'' in \emph{2018 IEEE International Solid-State Circuits
  Conference-(ISSCC)}.\hskip 1em plus 0.5em minus 0.4em\relax IEEE, 2018, pp.
  94--96.

\bibitem{hpmoravec_stereocart}
H.~P. Moravec, ``Obstacle avoidance and navigation in the real world by a
  seeing robot rover,'' Ph.D. dissertation, Stanford University, 1980.

\bibitem{goldberg2002stereo}
S.~B. Goldberg, M.~W. Maimone, and L.~Matthies, ``Stereo vision and rover
  navigation software for planetary exploration,'' in \emph{Proceedings, IEEE
  aerospace conference}, vol.~5.\hskip 1em plus 0.5em minus 0.4em\relax IEEE,
  2002, pp. 5--5.

\bibitem{kong2004method}
D.~Kong and H.~Tao, ``A method for learning matching errors for stereo
  computation.'' in \emph{BMVC}, vol.~1.\hskip 1em plus 0.5em minus 0.4em\relax
  Citeseer, 2004, p.~2.

\bibitem{zhang2007estimating}
L.~Zhang and S.~M. Seitz, ``Estimating optimal parameters for mrf stereo from a
  single image pair,'' \emph{IEEE Transactions on Pattern Analysis and Machine
  Intelligence}, vol.~29, no.~2, pp. 331--342, 2007.

\bibitem{spyropoulos2014learning}
A.~Spyropoulos, N.~Komodakis, and P.~Mordohai, ``Learning to detect ground
  control points for improving the accuracy of stereo matching,'' in
  \emph{Proceedings of the IEEE conference on computer vision and pattern
  recognition}, 2014, pp. 1621--1628.

\bibitem{zbontar2016stereo}
J.~Zbontar, Y.~LeCun, \emph{et~al.}, ``Stereo matching by training a
  convolutional neural network to compare image patches.'' \emph{J. Mach.
  Learn. Res.}, vol.~17, no.~1, pp. 2287--2318, 2016.

\bibitem{stein2006attenuating}
A.~N. Stein, A.~Huertas, and L.~Matthies, ``Attenuating stereo pixel-locking
  via affine window adaptation,'' in \emph{Proceedings 2006 IEEE International
  Conference on Robotics and Automation, 2006. ICRA 2006.}\hskip 1em plus 0.5em
  minus 0.4em\relax IEEE, 2006, pp. 914--921.

\bibitem{gordon2019depth}
A.~Gordon, H.~Li, R.~Jonschkowski, and A.~Angelova, ``Depth from videos in the
  wild: Unsupervised monocular depth learning from unknown cameras,'' in
  \emph{Proceedings of the IEEE/CVF International Conference on Computer
  Vision}, 2019, pp. 8977--8986.

\bibitem{godard2017unsupervised}
C.~Godard, O.~Mac~Aodha, and G.~J. Brostow, ``Unsupervised monocular depth
  estimation with left-right consistency,'' in \emph{Proceedings of the IEEE
  conference on computer vision and pattern recognition}, 2017, pp. 270--279.

\bibitem{wang2019unos}
Y.~Wang, P.~Wang, Z.~Yang, C.~Luo, Y.~Yang, and W.~Xu, ``Unos: Unified
  unsupervised optical-flow and stereo-depth estimation by watching videos,''
  in \emph{Proceedings of the IEEE/CVF Conference on Computer Vision and
  Pattern Recognition}, 2019, pp. 8071--8081.

\bibitem{smolyanskiy2018importance}
N.~Smolyanskiy, A.~Kamenev, and S.~Birchfield, ``On the importance of stereo
  for accurate depth estimation: An efficient semi-supervised deep neural
  network approach,'' in \emph{Proceedings of the IEEE Conference on Computer
  Vision and Pattern Recognition Workshops}, 2018, pp. 1007--1015.

\bibitem{scharstein_szeliski_hirschmuller}
\BIBentryALTinterwordspacing
D.~Scharstein, R.~Szeliski, and H.~Hirschmüller, ``Middlebury stereo
  benchmark.'' [Online]. Available: \url{https://vision.middlebury.edu/stereo/}
\BIBentrySTDinterwordspacing

\bibitem{kendall2017end}
A.~Kendall, H.~Martirosyan, S.~Dasgupta, P.~Henry, R.~Kennedy, A.~Bachrach, and
  A.~Bry, ``End-to-end learning of geometry and context for deep stereo
  regression,'' in \emph{Proceedings of the IEEE International Conference on
  Computer Vision}, 2017, pp. 66--75.

\bibitem{khamis2018stereonet}
S.~Khamis, S.~Fanello, C.~Rhemann, A.~Kowdle, J.~Valentin, and S.~Izadi,
  ``Stereonet: Guided hierarchical refinement for real-time edge-aware depth
  prediction,'' in \emph{Proceedings of the European Conference on Computer
  Vision (ECCV)}, 2018, pp. 573--590.

\bibitem{zhang2018activestereonet}
Y.~Zhang, S.~Khamis, C.~Rhemann, J.~Valentin, A.~Kowdle, V.~Tankovich,
  M.~Schoenberg, S.~Izadi, T.~Funkhouser, and S.~Fanello, ``Activestereonet:
  End-to-end self-supervised learning for active stereo systems,'' in
  \emph{Proceedings of the European Conference on Computer Vision (ECCV)},
  2018, pp. 784--801.

\bibitem{duggal2019deeppruner}
S.~Duggal, S.~Wang, W.-C. Ma, R.~Hu, and R.~Urtasun, ``Deeppruner: Learning
  efficient stereo matching via differentiable patchmatch,'' in
  \emph{Proceedings of the IEEE/CVF International Conference on Computer
  Vision}, 2019, pp. 4384--4393.

\bibitem{chang2018pyramid}
J.-R. Chang and Y.-S. Chen, ``Pyramid stereo matching network,'' in
  \emph{Proceedings of the IEEE Conference on Computer Vision and Pattern
  Recognition}, 2018, pp. 5410--5418.

\bibitem{cheng2020hierarchical}
X.~Cheng, Y.~Zhong, M.~Harandi, Y.~Dai, X.~Chang, T.~Drummond, H.~Li, and
  Z.~Ge, ``Hierarchical neural architecture search for deep stereo matching,''
  \emph{arXiv preprint arXiv:2010.13501}, 2020.

\bibitem{koinet}
P.~Wang, P.~Chen, Y.~Yuan, D.~Liu, Z.~Huang, X.~Hou, and G.~Cottrell,
  ``Understanding convolution for semantic segmentation,'' in \emph{2018 IEEE
  winter conference on applications of computer vision (WACV)}.\hskip 1em plus
  0.5em minus 0.4em\relax IEEE, 2018, pp. 1451--1460.

\bibitem{sceneflow}
N.~Mayer, E.~Ilg, P.~Hausser, P.~Fischer, D.~Cremers, A.~Dosovitskiy, and
  T.~Brox, ``A large dataset to train convolutional networks for disparity,
  optical flow, and scene flow estimation,'' in \emph{Proceedings of the IEEE
  conference on computer vision and pattern recognition}, 2016, pp. 4040--4048.

\bibitem{matchability}
J.~Zhang, Y.~Yao, Z.~Luo, S.~Li, T.~Shen, T.~Fang, and L.~Quan, ``Learning
  stereo matchability in disparity regression networks,'' in \emph{2020 25th
  International Conference on Pattern Recognition (ICPR)}.\hskip 1em plus 0.5em
  minus 0.4em\relax IEEE, 2021, pp. 1611--1618.

\bibitem{basler}
\BIBentryALTinterwordspacing
\emph{Basler Product Catalogue}, Basler, 2021. [Online]. Available:
  \url{https://www.baslerweb.com/en/products/cameras/area-scan-cameras/ace/aca2500-60uc}
\BIBentrySTDinterwordspacing

\bibitem{sunex}
\BIBentryALTinterwordspacing
\emph{Sunex DSL318 Specification Sheet}, Sunex, 2021. [Online]. Available:
  \url{http://www.optics-online.com/OOL/DSL/DSL318.PDF}
\BIBentrySTDinterwordspacing

\bibitem{replica19arxiv}
J.~Straub, T.~Whelan, L.~Ma, Y.~Chen, E.~Wijmans, S.~Green, J.~J. Engel,
  R.~Mur-Artal, C.~Ren, S.~Verma, A.~Clarkson, M.~Yan, B.~Budge, Y.~Yan,
  X.~Pan, J.~Yon, Y.~Zou, K.~Leon, N.~Carter, J.~Briales, T.~Gillingham,
  E.~Mueggler, L.~Pesqueira, M.~Savva, D.~Batra, H.~M. Strasdat, R.~D. Nardi,
  M.~Goesele, S.~Lovegrove, and R.~Newcombe, ``The {R}eplica dataset: A digital
  replica of indoor spaces,'' \emph{arXiv preprint arXiv:1906.05797}, 2019.

\bibitem{chen2020noise}
Y.~Chen, Z.~Lu, X.~Zhang, L.~Chen, and Q.~Liao, ``Noise-sampling cross entropy
  loss: Improving disparity regression via cost volume aware regularizer,'' in
  \emph{2020 IEEE International Conference on Image Processing (ICIP)}.\hskip
  1em plus 0.5em minus 0.4em\relax IEEE, 2020, pp. 2780--2784.

\bibitem{bajracharya2020mobile}
M.~Bajracharya, J.~Borders, D.~Helmick, T.~Kollar, M.~Laskey, J.~Leichty,
  J.~Ma, U.~Nagarajan, A.~Ochiai, J.~Petersen, \emph{et~al.}, ``A mobile
  manipulation system for one-shot teaching of complex tasks in homes,'' in
  \emph{2020 IEEE International Conference on Robotics and Automation
  (ICRA)}.\hskip 1em plus 0.5em minus 0.4em\relax IEEE, 2020, pp.
  11\,039--11\,045.

\bibitem{pytorch}
\BIBentryALTinterwordspacing
A.~Paszke, S.~Gross, F.~Massa, A.~Lerer, J.~Bradbury, G.~Chanan, T.~Killeen,
  Z.~Lin, N.~Gimelshein, L.~Antiga, A.~Desmaison, A.~Kopf, E.~Yang, Z.~DeVito,
  M.~Raison, A.~Tejani, S.~Chilamkurthy, B.~Steiner, L.~Fang, J.~Bai, and
  S.~Chintala, ``Pytorch: An imperative style, high-performance deep learning
  library,'' in \emph{Advances in Neural Information Processing Systems
  32}.\hskip 1em plus 0.5em minus 0.4em\relax Curran Associates, Inc., 2019,
  pp. 8024--8035. [Online]. Available:
  \url{http://papers.neurips.cc/paper/9015-pytorch-an-imperative-style-high-performance-deep-learning-library.pdf}
\BIBentrySTDinterwordspacing

\bibitem{middlebury}
D.~Scharstein, H.~Hirschm{\"u}ller, Y.~Kitajima, G.~Krathwohl,
  N.~Ne{\v{s}}i{\'c}, X.~Wang, and P.~Westling, ``High-resolution stereo
  datasets with subpixel-accurate ground truth,'' in \emph{German conference on
  pattern recognition}.\hskip 1em plus 0.5em minus 0.4em\relax Springer, 2014,
  pp. 31--42.

\bibitem{xu2020aanet}
H.~Xu and J.~Zhang, ``Aanet: Adaptive aggregation network for efficient stereo
  matching,'' in \emph{Proceedings of the IEEE/CVF Conference on Computer
  Vision and Pattern Recognition}, 2020, pp. 1959--1968.

\bibitem{iResNet}
Z.~Liang, Y.~Feng, Y.~Guo, H.~Liu, L.~Qiao, W.~Chen, L.~Zhou, and J.~Zhang,
  ``Learning deep correspondence through prior and posterior feature
  constancy,'' \emph{arXiv preprint arXiv:1712.01039}, vol.~7, no.~8, 2017.

\bibitem{hitnet}
V.~Tankovich, C.~Hane, Y.~Zhang, A.~Kowdle, S.~Fanello, and S.~Bouaziz,
  ``Hitnet: Hierarchical iterative tile refinement network for real-time stereo
  matching,'' in \emph{Proceedings of the IEEE/CVF Conference on Computer
  Vision and Pattern Recognition}, 2021, pp. 14\,362--14\,372.

\bibitem{hsmnet}
G.~Yang, J.~Manela, M.~Happold, and D.~Ramanan, ``Hierarchical deep stereo
  matching on high-resolution images,'' in \emph{Proceedings of the IEEE/CVF
  Conference on Computer Vision and Pattern Recognition}, 2019, pp. 5515--5524.

\bibitem{adastereo}
X.~Song, G.~Yang, X.~Zhu, H.~Zhou, Z.~Wang, and J.~Shi, ``Adastereo: a simple
  and efficient approach for adaptive stereo matching,'' in \emph{Proceedings
  of the IEEE/CVF Conference on Computer Vision and Pattern Recognition}, 2021,
  pp. 10\,328--10\,337.

\end{thebibliography}
\end{document}